\pdfoutput=1

\documentclass[11pt]{article}

\usepackage{ACL2023}

\usepackage{times}
\usepackage{latexsym}
\usepackage{amsmath}
\usepackage{graphicx}

\usepackage[T1]{fontenc}

\usepackage[utf8]{inputenc}

\usepackage{microtype}

\usepackage{inconsolata}
\usepackage{enumitem}
\title{Conversational Recommendation as Retrieval: A Simple, Strong Baseline}

\author{Raghav Gupta, Renat Aksitov, Samrat Phatale, Simral Chaudhary, Harrison Lee, Abhinav Rastogi \\
  Google Research \\
  \texttt{\{raghavgupta, raksitov, samratph, simral, harrisonlee, abhirast\}@google.com} \\}

\begin{document}
\maketitle
\begin{abstract}
Conversational recommendation systems (CRS) aim to recommend suitable items to users through natural language conversation. However, most CRS approaches do not effectively utilize the signal provided by these conversations. They rely heavily on explicit external knowledge e.g., knowledge graphs to augment the models' understanding of the items and attributes, which is quite hard to scale. To alleviate this, we propose an alternative information retrieval (IR)-styled approach to the CRS item recommendation task, where we represent conversations as queries and items as documents to be retrieved. We expand the document representation used for retrieval with conversations from the training set. With a simple BM25-based retriever, we show that our task formulation compares favorably with much more complex baselines using complex external knowledge on a popular CRS benchmark. We demonstrate further improvements using user-centric modeling and data augmentation to counter the cold start problem for CRSs.
\end{abstract}

\section{Introduction}

Recommendation systems have become ubiquitous in recent years given the explosion in massive item catalogues across applications. In general, a recommendation system learns user preference from historical user-item interactions, and then recommends items of user's preference. In contrast, CRSs directly extract user preferences from live dialog history to precisely address the users’ needs. An example dialogue from the popular ReDial benchmark \citep{redial} for CRSs is shown in Table \ref{tab:redial_ex}: the CRS' task is to recommend items (in this case, {\color{blue}movies}) based on the user's indicated preference.

\begin{table}[t!]
    \centering
    \setlength{\tabcolsep}{3pt}
    \scalebox{0.75}{
    \begin{tabular}{|l|p{8cm}|}
    \hline\hline
\textbf{Role} & \textbf{Message} \\\hline\hline
User & Hello! I am looking for some movies.\\\hline
Agent & What kinds of movie do you like? I like {\color{red}animated} movies such as {\color{blue}Frozen (2013)}.\\\hline
Rec. item & \textcolor{blue}{Frozen (2013)}\\\hline
User & I do not like {\color{red}animated films}. I would love to see a movie like {\color{blue}Pretty Woman (1990)} starring {\color{red}Julia Roberts}. Know any that are similar?\\\hline
Agent & \textcolor{blue}{Pretty Woman (1990)} was a good one. If you are in it for {\color{red}Julia Roberts} you can try {\color{blue}Runaway Bride (1999)}.\\\hline
Rec. item & \textcolor{blue}{Runaway Bride (1999)}\\\hline
    \end{tabular}}
    \caption{An example dialogue from ReDial. The items to recommend are in blue, with their inferred attributes in red. The ground truth recommended items for agent utterances are also shown.}
    \label{tab:redial_ex}
\end{table}

Generally, a CRS integrates two modules: a \textbf{dialogue module} which generates natural language responses to interact with users, and a \textbf{recommendation module} which recommends desirable items to users using the dialog context and external knowledge. We focus on the latter module in this work: we posit that once the correct item to recommend is identified, newer pretrained language models (PLMs) can easily generate fluent agent responses.

It is notable that the conversational context provides sufficient signal to make good recommendations \citep{yang2021improving}. E.g., in Table \ref{tab:redial_ex}, attributes about the items to recommend (e.g., genre and cast, in {\color{red}red}) provide potentially sufficient information to the model to recommend relevant items.

Most approaches to CRS rely heavily on external knowledge sources, such as knowledge graphs (KGs) and reviews \citep{lu-etal-2021-revcore}. Such approaches require specific sub-modules to encode information from these sources like graph neural networks \citep{kipf2016semi}, which are hard to scale with catalog additions. Existing approaches require either re-training the entire system when the KG structure
changes \citep{dettmers2018convolutional} or adding complex
architectures on top to adapt \citep{wu2022efficiently}. Newer approaches utilize PLMs \citep{radford2018improving, lewis2020bart}, but they often encode item information in model parameters, making it hard to scale to new items without retraining.

Looking for a fast, more scalable approach, we re-formulate the item recommendation task for CRSs as an information retrieval (IR) task, with recommendation-seeking conversations as queries and items to recommend as documents. The document content for retrieval is constructed using plain text metadata for the item paired with conversations where the said item is recommended, in order to enhance semantic overlap between the queries which are themselves conversations.

We apply a standard non-parametric retrieval baseline - BM25 - to this task and show that the resulting model is fast and extensible without requiring complex external knowledge or architectures, while presenting improvements over more complex item recommendation baselines. Our contributions are summarized as follows:

\begin{itemize}[leftmargin=*]
\itemsep0em
    \item We present an alternate formulation of the CRS recommendation task as a retrieval task.
    \item We apply BM25 to this task, resulting in a simple, strong model with little training time and reduced reliance on external knowledge.
    \item We further improve the model using user-centric modeling, show that the model is extensible to new items without retraining, and demonstrate a simple data augmentation method that alleviates the cold start problem for CRSs.
\end{itemize}



\section{Related Work}
\label{sec:relwork}
\noindent Conversational recommendation systems constitute an emerging research area, helped by datasets like
REDIAL \citep{redial}, TG-REDIAL \citep{zhou2020topicguided},
INSPIRED \citep{hayati2020inspired}, DuRecDial \citep{liu2020towards, liu2021durecdial}, and CPCD \citep{Chaganty2023BeyondSI}. We next describe the recommender module architectures of CRS baselines.

ReDial \citep{redial} uses an autoencoder to generate recommendations. CRSs commonly use knowledge graphs (KGs) for better understanding of the item catalog: DBpedia \citep{auer2007dbpedia} is a popular choice of KG. KBRD \citep{kbrd} uses item-oriented KGs, while KGSF \citep{zhou2020improving} further incorporates a word-based KG \citep{speer2017conceptnet}. CR-Walker \citep{ma2021cr} performs tree-structured reasoning on the KG, CRFR \citep{zhou2021crfr} does reinforcement learning and multi-hop reasoning on the KG. UniCRS \citep{wang2022towards} uses knowledge-added prompt tuning with and KG \& a fixed PLM. Some methods also incorporate user information: COLA \citep{lin2022cola} uses collaborative filtering to build a user-item graph, and \citep{li2022user} aims to find lookalike users for user-aware predictions.

Eschewing KGs, MESE \citep{yang-etal-2022-improving} trains an item encoder to convert flat item metadata to embeddings then used by a PLM, and TSCR \cite{zou2022improving} trains a transformer with a Cloze task modified for recommendations. Most above approaches, however, either rely on complex models with KGs and/or need to be retrained for new items, which is very frequent in present-day item catalogs.

\section{Model}

We formally define the item recommendation task, followed by our retrieval framework, details of the BM25 retrieval model used, and finally our user-aware recommendation method on top of BM25.


\subsection{Conversational Item Recommendation}
A CRS allows the user to retrieve relevant items from an item catalog $V = \{v_1, v_2\cdots v_N\}$ through dialog. In a conversation, let $a$ be an agent response containing an item(s) from $V$ recommended to the user. Let $d_t = \{u_{1}, u_{2}, \cdots\ u_{t}\}$ be the $t$ turns of the conversation context preceding $a$, where each turn can be spoken by the user or the agent.

We model the recommendation task as masked item prediction, similar to \citet{zou2022improving}. For each agent response $a$ where an item $v_i \in V$ is recommended, we mask the mention of $v_i$ in $a$ i.e. replace it with the special token \texttt{[REC]}, yielding the masked agent response $a'$. We now create training examples with input $q = d_t \oplus a'$ and ground truth $v_i$ ($\oplus$ denotes string concatenation).

We define $Q^{train}$ and $Q^{test}$ as the set of all conversational contexts $q = d_t\oplus a'$ with an item to predict, from the training and test sets respectively. For each item $v_i$, we also define $\boldsymbol{Q^{train}_{v_i}} \subset Q^{train}$ as the set of all conversational contexts in $Q^{train}$ where $v_i$ is the ground truth item to recommend.

\subsection{Item Recommendation as Retrieval}
Information retrieval (IR) systems are aimed at recommending documents to users based on the relevance of the document's content to the user query. We reformulate masked item prediction as a retrieval task with $Q^{train}$ or $Q^{test}$ as the set of queries to calculate relevance to, and $V$ as the set of items/documents to recommend from.

To match a query $q\in Q^{test}$ to a document/item $v_i \in V$, we define the document's content using two sources: \textbf{metadata} in plaintext about item $v_i$, and $\boldsymbol{Q^{train}_{v_i}}$ i.e. all conversational contexts from the training set where $v_i$ is the recommended item, concatenated together, similar to document expansion \citep{nogueira2019document}. Our motivation for adding $Q^{train}_{v_i}$ to the document representation is that it is easier to match queries (which are conversations) to conversations instead of plain metadata since conversations can be sparse in meaningful keywords. For an item $v_i$ we create a document as:
\begin{equation}
  Doc(v_i) = Metadata(v_i) \oplus Q_{v_i}
\end{equation}

For test set prediction, we can now apply retrieval to recommend the most relevant document $Doc(v_i), v_i\in V$, for each test set query $q \in Q^{test}$.

\subsection{Retrieval Model: BM25}
\label{sec:bm25}
BM25 \citep{bm25} is a commonly used sparse, bag-of-words ranking function. It produces a similarity score for a given document, $doc$ and a query, $q$, by matching keywords efficiently with an inverted index of the set of documents. Briefly, for each keyword in each document, we compute and store their term frequencies (TF) and inverse document frequencies (IDF) in an index. For an input query, we compute a match score for each query keyword with each document using a function of TF and IDF, and sum this score over all keywords in the query. This yields a similarity score for the query with each document, which is used to rank the documents for relevance to the query.





\subsection{User Selection}
Our IR formulation also gives us a simple way to incorporate user information for item recommendation. Let $U = \{u_1, u_2\ldots u_J\}$ be the set of all users in the dataset. Each conversation context in $Q^{train}$ be associated with a user $u_j\in U$. We use a simple algorithm for user-aware recommendations:
\begin{itemize}[leftmargin=*]
\itemsep0em
    \item For each user $u\in U$, we obtain the set of items they like based on conversations in $Q^{train}$, and also construct a unique BM25 index for each user $u_j$ using only conversations associated with $u_j$.
    \item For a test set query $q\in Q^{test}$, we identify movies liked by the seeker in the current $q$, and use it to find the $M$ most similar users in the training set.
    \item We now compute and add up similarity scores for the query with all documents based on the per-user BM25 indices for these $M$ selected users.
    \item Finally, we linearly combine these user-specific similarity scores per document with the similarity scores from the BM25 index in Section \ref{sec:bm25}, and use these combined scores to rank all documents.
\end{itemize}

\section{Experiments}
\subsection{Dataset and Evaluation}
ReDial \citep{redial} is a popular benchmark of annotated dialogues where a seeker requests movie suggestions from an agent. Figure \ref{tab:redial_ex} shows an example. It contains 956 users, 51,699 movie mentions, 10,006 dialogues, and 182,150 utterances.

For evaluation, we reuse Recall@$k$ (or R@$k$) as our evaluation metric for ReDial from prior work. It evaluates whether the target human-recommended item appears in the top-$k$ items produced by the recommendation system. We compare against baselines introduced in Section \ref{sec:relwork}.

\subsection{Training}

For movie recommendations, we extract metadata from \textit{IMDb.com} to populate $Metadata(v_i)$ for movies $v_i \in V$, which includes the movie's brief plot and names of the director and actors.

Parameters $k_1$ and $b$ for BM25 are set to 1.6 and 0.7 respectively. For user selection, we select the $K=5$ most similar users, and linearly combine the user-specific BM25 scores with the overall BM25 scores with a coefficient of 0.05 on the former. Constructing the BM25 index on the ReDial training set and inference on the test set took \textasciitilde5 minutes on a CPU (+10 minutes for the user selection method). Alongside BM25 with and without user selection, we also experiment with a BM25 variant without metadata i.e. using only past conversation contexts as the document content for a movie/item.

\begin{table}[t!]
    \centering
    \setlength{\tabcolsep}{3pt}
    \scalebox{0.9}{
    \begin{tabular}{l|c|c|c}\hline
    \textbf{Model} & \textbf{R@1} & \textbf{R@10} & \textbf{R@50}\\\hline
    ReDial \citep{redial} & 2.3 & 12.9 & 28.7\\
KBRD* \citep{kbrd} & 3.0 & 16.4 & 33.8\\
KGSF* \citep{zhou2020improving} & 3.9 & 18.3 & 37.8\\
CR-Walker* \citep{ma2021cr} & 4.0 & 18.7 & 37.6\\
CRFR* \citep{zhou2021crfr} & 4.0 & 20.2 & 39.9\\
COLA* \citep{lin2022cola} & 4.8 & 22.1 & 42.6\\
UniCRS* \citep{wang2022towards} & 5.1 & 22.4 & 42.8\\
MESE$\dagger$ \citep{yang2021improving} & 5.6 & 25.6 & 45.5\\
TSCR* \citep{zou2022improving} & 7.2 & 25.7 & 44.7\\
\hline
BM25 w/o Metadata & 4.8 &	19.5&	37.4\\
BM25$\dagger$ & 5.2 &	20.5 &	38.5\\
BM25 + User Selection$\dagger$ & 5.3 &21.1 &	38.7\\
\hline
    \end{tabular}
    }
    \caption{Item recommendation results on the ReDial benchmark. Our BM25-based models outperform many baselines despite being much, lighter and not using complex KGs. * denotes models using DBPedia KG, $\dagger$ denotes models using plaintext IMDb metadata.}
    \label{tab:results}
\end{table}

\section{Results}
Table \ref{tab:results} shows $R@\{1, 10, 50\}$ on ReDial for the baselines and our models. Our BM25-based models perform strongly, outperforming many baselines which use complex KGs and/or complex model architectures e.g., tree-structured reasoning and reinforcement learning. Improvement is most visible on $R@1$ and less so on $R@50$. Our fairest comparison would be with \textbf{MESE}, which uses the exact same data (plaintext metadata + dialog context): our best model achieves $95\%$ of its $R@1$ and 85\% of its $R@50$ with a far faster and simpler model. We point out that all baselines except TSCR are jointly optimized for both the item recommendation and response generation tasks.

A surprising result is \textbf{BM25 w/o Metadata} doing better than many baselines, without using any external knowledge whatsoever, in contrast to all other baselines except \textbf{ReDial}. This indicates that prior conversations indeed contain sufficient signal for good conversational item recommendation.

Our simple \textbf{user selection} raises recall by 1-3\% across thresholds, with more potential gains from better user-centric modeling \citep{li2022user}.

\section{Cold Start and Data Augmentation}
Conversational recommenders often suffer from the \textbf{cold start problem}: it is difficult for a new item i.e. not seen during training, to be recommended, since not much is known about it beyond metadata. 

\begin{figure}[t!]
    \centering
    \includegraphics[width=\linewidth]{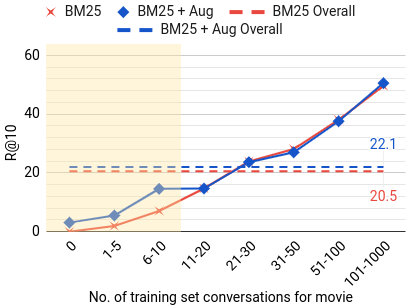}
    \caption{Impact of data augmentation on $R@10$. The shaded area represents the set of movies for which data augmentation was performed.}
    \label{fig:data_aug1}
\end{figure}

Our model is not immune to this problem. The red lines in Figure \ref{fig:data_aug1} show $R@10$ values for the BM25 model for different sets of movies in ReDial based on how many times they are seen in the training set: the model never or rarely recommends movies with 10 or fewer occurrences in training.

To counteract this, we perform \textbf{data augmentation} using few-shot prompting \citep{liu2023pre}. In particular, we randomly select 6 conversations from ReDial's training set, use them to prompt a PaLM 2-L model \citep{anil2023palm}, and generate up to 20 dialogues per movie. We do this only for movies seen 10 or fewer times during training, since the model does the worst on these.

\begin{figure}
    \centering
    \includegraphics[width=\linewidth]{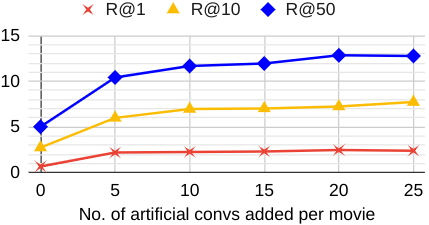}
    \caption{Recall for the BM25 model with varying amounts of augmented conversations.}
    \label{fig:data_aug2}
\end{figure}
Figure \ref{fig:data_aug1}'s blue curve shows notably improved $R@10$ for the movies for which data was augmented, without hurting $R@10$ for more frequent movies. Overall $R@10$ also improves by \textasciitilde8\% using just $\leq$ 20 artificial dialogues per movie. Further combining augmentation with user selection lifts $\boldsymbol{R@1}$ to \textbf{5.9}, $\boldsymbol{R@10}$ to \textbf{22.3}, and $\boldsymbol{R@50}$ to \textbf{40.7}.

Figure \ref{fig:data_aug2} plots recall for BM25 model with the number of artificial dialogues added for low-frequency movies. Based on this plot, we opted to generate at most 20 conversations per movie.

\section{Conclusion}
We present a retrieval-based formulation of the item recommendation task, used in building CRSs, by modeling conversations as queries and items to recommend as documents. We augment the item representation with conversations recommending that item, therefore the retrieval task reduces to matching conversations to conversations, which is feasible and effective. Using BM25-based retrieval with this IR task results in a model that is very fast and inexpensive to train/build (\textasciitilde5 min on CPU) while being flexible to add-ons such as user selection. We also show that new items can be seamlessly added without retraining the entire model, and that simple data augmentation with as few as 20 conversations counters the cold start problem for a new item: far fewer than most neural network finetuning-based methods would need.

\bibliography{anthology,custom}
\bibliographystyle{acl_natbib}




\end{document}